\pdfoutput=1

\documentclass[11pt]{article}

\usepackage[final]{coling}

\usepackage{times}
\usepackage{latexsym}

\usepackage[T1]{fontenc}

\usepackage[utf8]{inputenc}

\usepackage{microtype}

\usepackage{inconsolata}

\usepackage{graphicx}

\title{ESURF: Simple and Effective EDU Segmentation}

\author{Mohammedreza Sediqin \\
  Illinois Institute of Technology\\
  \texttt{msediqin@hawk.iit.edu} \\\And
  Shlomo Engelson Argamon \\
  Touro University \\
  \texttt{shlomo.argamon@touro.edu} \\}

\begin{document}
\maketitle

\begin{abstract}
Segmenting text into Elemental Discourse Units (EDUs) is a fundamental task in discourse parsing. We present a new simple method for identifying EDU boundaries, and hence segmenting them, based on lexical and character n-gram features, using random forest classification. We show that the method, despite its simplicity, outperforms other methods both for segmentation and within a state of the art discourse parser. This indicates the importance of such features for identifying basic discourse elements, pointing towards potentially more training-efficient methods for discourse analysis.

\end{abstract}

\section{Introduction}
A fundamental task in natural language understanding is analyzing the overall structure of a text, so that logical and coherence relations between text units are revealed.  Rhetorical Structure Theory (RST)~\cite{mann1988rhetorical} is a well-acccepted theoretical framework for the task within the NLP community \cite{ kobayashi-AAAI2020}. RST structures a text as a tree, where the basic building blocks (leaf node) are called Elementary Discourse Units (EDUs). Discourse parsing in RST is the task of automatically constructing this hierarchical tree by identifying the EDUs and then building a parse tree by connecting adjacent EDUs and composite discourse units. Relations between adjacent units are labeled with different rhetorical relations, which are mostly assymetrical, with one unit designated the the nucleus and the other as the subordinate satellite. Parsing is generally done using a shift-reduce parser, which builds the tree incrementally by scoring transition actions \cite{yu-iccl18, mabona2019neural}. Recently, neural network models have achieved state-of-the-art performance in this task by leveraging sophisticated neural modules \cite{zhang2020top}).


Rhetorical Structure Theory (RST) offers a robust approach for discourse analysis by constructing a rhetorical structure tree that captures the relationships between text elements, enhancing performance in various tasks. Although previous research efforts have advanced machine learning methods for discourse segmentation and parsing, these often rely on lexical and syntactic clues, hand-crafted features, and syntactic parse trees, and use gold-standard segmentation for training and evaluation \cite{yu2022rst, feng2014two, ali2023fuzzy}. This coherence structure is essential for applications such as  text summarization, and sentiment analysis. RST-based analysis significantly improves discourse understanding and contributes to more effective NLP applications \cite{nguyen2021rst,liu2021dmrst}.

Despite the successes of contextualized pre-trained language models (PLMs) like XLNet \cite{yang2020xlnetgeneralizedautoregressivepretraining} in RST discourse parsing, challenges remain due to data insufficiency, reliance on lexical and syntactic clues, and inconsistencies between EDU-level parsing and sentence-level contextual modeling, as well as dependence on gold-standard segmentation for training. These issues, particularly the reliance on hand-crafted features and parse trees, have made EDU segmentation a significant bottleneck. In this paper, we propose a novel method for EDU segmentation which gives state-of-the-art (SOTA) performance, showing that local lexical and morphological cues can do most of the work. 

 We conduct experiments using the RST Discourse Treebank (RST-DT) and CNN/Daily Mail. First, we derive EDU segmentation and evaluate it with various classifiers, including transformers. We then test our proposed EDU identification method using a transition-based neural RST parser \cite{yu2022rst}. Our results demonstrate improvements in EDU identification and RST parsing, with our model outperforming others and improving automated RST parsing techniques.

\section{Related work}

Historically, discourse processing using RST has been approached as a parsing task, using transition-based or chart parsers \cite{luong2015multi,dai2019regularization,li2022survey}. In recent years, performance has been improved over earlier methods by incorporating statistical models for predicting nuclearity and relation types between discourse units \cite{yu2018transition,kobayashi-AAAI2020,zhang2020top,guz2020coreference,koto-etal-2021-top}. Such neural approaches now dominate, but many still incorporate hand-crafted features for better performance. 
%
Seq2Seq models have also been applied to both sentence and document-level parsing \cite{liu2019dynamic,luong2015multi, dai2019regularization}. 

A key component of discourse parsing is identifying the Elementary Discourse Units (EDUs) defined as smallest text spans. 
Early methods relied on handcrafted features and syntactic clues \cite{mann1988rhetorical, lan2013leveraging}. 
Recent neural models like BERT and XLNet have advanced EDU segmentation and discourse coherence \cite{zhang-elmo-etal-2021-adversarial, yu2022rst}.

Some recent work has focused on developing better parsing methods independent of EDU segmentation, by using a gold-standard segmentation for training and evaluating RST parsers, and employing top-down approaches with sequence labeling for RST parsing \cite{nguyen2021rst,mabona2019neural,koto-etal-2021-top}. Such work gives strong baselines for parsing using different methods.

As noted above, we focus on the core subtask of EDU segmentation, and will evaluate our method both for segmentation accuracy and for its effect on parsing accuracy.


\begin{figure*}[bt]
\vspace{-40pt}
\centering
  \includegraphics[width=0.7\linewidth]{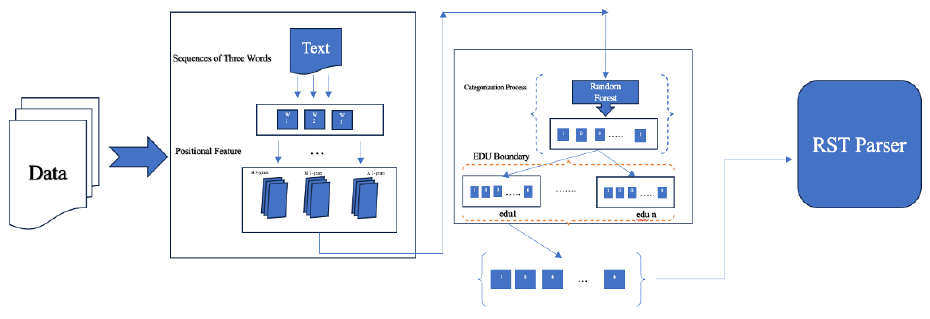} \vspace{-10pt}
  \caption {Architecture diagram for ESURF\label{fig:architecture}}
\end{figure*}

\begin{table*}[tb]
  \centering
  \scriptsize
  \begin{tabular}{lcccccccc}
    \hline
    \textbf{Model} & \textbf{Acc. (CNN)} & \textbf{Prec. (CNN)} & \textbf{Rec. (CNN)} & \textbf{F1 (CNN)} & \textbf{Acc. (RST-DT)} & \textbf{Prec. (RST-DT)} & \textbf{Rec. (RST-DT)} & \textbf{F1 (RST-DT)} \\
    \hline
    SVM & 0.851 & 0.843 & 0.858 & 0.850 & 0.889 & 0.885 & 0.881 & 0.886 \\
    CRF & 0.893 & 0.878 & 0.852 & 0.865 & 0.927 & 0.903 & 0.875 & 0.885 \\
    Gradient Boosted & 0.871 & 0.875 & 0.866 & 0.873 & 0.890 & 0.904 & 0.891 & 0.885 \\
    \textbf{ESURF} & \textbf{0.912} & \textbf{0.894} & \textbf{0.921} & \textbf{0.907} & \textbf{0.950} & \textbf{0.935} & \textbf{0.979} & \textbf{0.958} \\
    BERT & 0.903 & 0.888 & 0.910 & 0.899 & 0.943 & 0.909 & 0.953 & 0.947 \\
    XLNet & 0.472 & 0.256 & 0.470 & 0.335 & 0.496 & 0.253 & 0.491 & 0.330 \\
    \hline
  \end{tabular}
  \caption{Classifier performance for EDU identification on the CNN/Daily Mail and RST-DT datasets.  ESURF achieved the highest metrics for both.
  \label{tab:classifier_seg}}
\end{table*}

\section{EDU Segmentation Using Random Forests (ESURF)}
The significance of EDUs in RST parsing is crucial due to their fundamental role in understanding discourse structures. 
EDUs represent the smallest coherent ``thought units'' within a text, and are the parts of which the overall discourse structure is composed. 
Hence, accurate segmentation and identification of EDUs is essential for accurate analysis of rhetorical structure \cite{yu2018transition,lin2023natural}. 


We present here a comparatively simple, but quite effective, method for EDU segmentation, which we call \textit{EDU Segmentation Using Random Forests (ESURF)}. 
ESURF formulates the problem of EDU segmentation as a classification problem.
The system considers every nine-token\footnote{Some subsequences are shorter, as we do not consider sequences that cross sentence boundaries.} subsequence of the text $(t_{i-3},t_{i-2},t_{i-1},t_{i},...,t_{i+5})$,as a possible context for an EDU boundary, giving three tokens before and six tokens after the candidate boundary (immediately preceding $t_i$).
The input features for classification are the individual tokens $t_k$ given per their position in the context window, marked as \textit{\textbf{B}efore} ($t_{i-3},t_{i-2},t_{i-1}$), \textit{\textbf{L}eading} ($t_{i},t_{i+1},t_{i+2}$), or \textit{\textbf{C}ontinuing} ($t_{i+3},t_{i+4},t_{i+5}$) the candidate EDU.
To account in a simple way for morphology, we also added as potential features character subsequences of the tokens, similarly marked as B, L, or C. 
These were filtered to keep only the character subsequences that appeared in multiple corpus texts, but not in most of them, as a simple measure of informativeness.

We train a random-forest classifier on these examples using these features. The classifier is then used to classify all candidate EDU boundaries in new text, processing each 9-token window as above for classification. Each sequence of tokens between boundaries classified as positive (i.e., as an EDU boundary) is identified as an EDU. Figure~\ref{fig:architecture} gives a schematic diagram of the overall system.


\section{Results}

We perform two sets of evaluations. First, we compare the performance of ESURF against other classification methods and other EDU segmentation methods from the literature on the task of EDU segmentation. Next, we evaluate the effect on RST parsing performance of using ESURF for segmentation, as compared with the methods used in various recent RST parsing systems.

\subsection{Datasets}
In all of our experiments, we use the RST Discourse Treebank (RST-DT) dataset \cite{carlson2002rst}, which is standard in the area of RST parsing. RST-DT consists of 347 training articles and 38 test articles annotated with full RST discourse structures.
Additionally, we also evaluate ESURF on the CNN/Daily Mail dataset \cite{nallapati2016abstractive}, which includes over 300,000 articles.

\subsection{ESURF on CNN/DailyMail and RST-DT Datasets}
We evaluate ESURF against various classifiers on the task of classifying sections as EDUs or non-EDUs. This comparison includes CRF (as a classifier), BERT, and XLNet, using the CNN/Daily Mail and RST-DT datasets. The evaluation is performed on a similarly sized subset of data points (50\% positive / 50\% negative) with preprocessing consistent with our previous approach. As shown in Table~\ref{tab:classifier_seg}, ESURF outperforms the other models in accuracy, precision, recall, and F1 score. 

For the CNN/Daily Mail dataset, ESURF achieves an accuracy of 91.2\% and an F1-score of 90.7\%, outperforming BERT and the other segmenters in this comparison.

These results show that, despite its simplicity, ESURF is highly effectiveness in EDU segmentation, indicating the centrality of lexical and morphological context as cues for discourse segmentation.
We further evaluate our model against several established discourse segmenters, widely recognized as baselines in EDU segmentation studies. Shown in Table~\ref{tab:seg_performance}, this comparison includes JCN, which uses a Logistic Regression model with features from sentence context, combining syntactic tree structures and statistical estimates. We also assess CRF and WLY, which apply sequence labeling and a BiLSTM-CRF framework, and HILDA (HIL) and SPADE (SP), which utilize statistical models integrating syntactic and lexical information to identify discourse boundaries and build sentence-level discourse trees.

Results on the RST-DT test set show that our model, ESURF, again outperforms the other methods. Specifically, while the Joint Model achieves an F1-score of 95.5\%, ESURF improves upon this with an F1-score of 95.8\%. This improvement in performance suggests a potential increase in RST parsing accuracy. 
\begin{table}[t]

  \centering
  \scriptsize
  \vspace{-10pt}
  \begin{tabular}{lccc}
    \hline
    \textbf{Model} & \textbf{Precision} & \textbf{Recall} & \textbf{F1 Score} \\
    \hline
    Hill \cite{hernault2010hilda} & 0.779 & 0.706 & 0.741 \\
    SP\cite{soricut2003sentence} & 0.838 & 0.868 & 0.852 \\
   CRF
   \cite{feng2014linear} & 0.903 & 0.918 & 0.905 \\
    JCN 
    \cite{joty2012novel} & 0.880 & 0.923 & 0.901 \\
    WYL
    \cite{wang2018toward}& 0.924 & 0.944 & 0.932 \\
    F\&R \cite{fisher2007utility} & 0.913 & 0.897 & 0.905 \\
    Joint Model \cite{lin2023natural} & 0.933 & 0.978 & 0.955 \\
    \textbf{ESURF} & \textbf{0.935} & \textbf{0.979} & \textbf{0.958} \\
    \hline
  \end{tabular}
  \vspace{-5pt}
  \caption{Performance comparison of various established EDU segmentation methods for RST-DT. 
  \label{tab:seg_performance}}
\end{table}

\subsection{ RST Parsing Using ESURF}
We evaluate the impact of various parsing methods, including our EDU segmentation model, by using it in a state-of-the-art RST parser. Our analysis demonstrates that enhancing EDU segmentation significantly improves discourse parsing performance. we utilize a Shift-reduce transition-based neural RST parser to assess the effectiveness of our EDU segmentation method. We employ the parser developed by\cite{yu2022rst}, which leverages neural RST parsing to produce action sequences from EDU representations. 

For evaluation, we adopt the framework proposed by \cite{morey2017much}, which evaluates performance using microaveraged $F_1$ for using four scoring metrics: Span (tree structure without labels), Nuclearity (structure with just nuclearity labels), Relation (structure with just relation labels), and Full (structure with both nuclearity and relation labels). In Table~\ref{tab:parser_performance}, we compare our results against various leading RST parsers. 

Our segmentation method ESURF improves the performance of the transition-based RST parser from \cite{yu2022rst} by approximately 0.5\% in Span, more than 1\% in Relation, and 0.3\% in Full metrics. \cite{yu2022rst} reimplemented the EDU segmenter from Muller \cite{muller2019tony} for segmenting large-scale unlabeled texts. This improvement highlights how improved EDU segmentation can enhance RST parsing performance. 
\begin{table}[t]
 \centering
 \scriptsize
 \vspace{-10pt}
  \begin{tabular}{lcccc}
    \hline
    \textbf{Model} & \textbf{S } & \textbf{N } & \textbf{R } & \textbf{F } \\
    \hline
    
    \cite{yu2022rst} & 0.764 &\textbf {0.661} & 0.545 & 0.535 \\
    \cite{zhang2021adversarial} & 0.763 & 0.655 & 0.556 & 0.538 \\
    \cite{yu2018transition} & 0.714 & 0.603 & 0.492 & 0.481 \\
    \cite{zhang2020top} & 0.672 & 0.555 & 0.453 & 0.443 \\
    \cite{nguyen2021rst} & 0.743 & 0.643 & 0.516 & 0.502 \\
    \cite{koto2021top} & 0.731 & 0.623 & 0.515 & 0.503 \\
   \textbf{\cite{yu2022rst}+ESURF} & \textbf{0.768} & {0.659} & \textbf{0.558}& \textbf{0.538} \\
    \hline
    
  \end{tabular} 
  \caption{Performance comparison of various RST Parser with metrics S , N , R , and F.
  \label{tab:parser_performance}}
\end{table}

\section{Conclusion and Future Work}
We demonstrate that our method ESURF, despite its simplicity, achieves SOTA performance on EDU segmentation, and also improves RST parsing performance to SOTA as well. This shows that lexical and morphological context give strong cues for identifying basic discourse structure constituents.


For future work, we plan to apply ESURF to large unlabeled datasets like the GUM corpus to give a semi-supervised method for EDU segmentation, potentially improving training efficiency for application to lower-resource languages. 

\section{Limitation}
A key limitation of this study is its evaluation on only the RST-DT and CNN/DailyMail datasets, omitting other discourse benchmarks like the Penn Discourse Treebank (PDTB) and the GUM corpus. Additionally, the focus is solely on sentence-level discourse parsing, and assumes accurate sentence segmentation. Future work should relax this assumption by using an automated sentence segmenter, explore different features and parameter settings to evaluate their impact, and evaluate/compare methods on a broader range of datasets.
\bibliography{ref}

\end{document}